\documentclass[letterpaper, 10 pt, conference]{ieeeconf}  % Comment this line out if you need a4paper

\IEEEoverridecommandlockouts                              % This command is only needed if 
\usepackage{dblfloatfix} 
\usepackage[utf8]{inputenc} % allow utf-8 input
\usepackage[T1]{fontenc}    % use 8-bit T1 fonts
\usepackage{hyperref}       % hyperlinks
\usepackage{url}            % simple URL typesetting
\usepackage{booktabs}       % professional-quality tables
\usepackage{amsfonts}       % blackboard math symbols
\usepackage{nicefrac}       % compact symbols for 1/2, etc.
\usepackage{microtype}      % microtypography
\usepackage{amsmath,amsthm}
\usepackage{subcaption}
\usepackage{wrapfig,graphicx}
\usepackage{color}
\usepackage{comment}
\usepackage{mathtools}
\usepackage{float}
\usepackage[linesnumbered,ruled,noend]{algorithm2e}
\usepackage[font=small,labelfont=bf]{caption}

\usepackage{graphics} % for pdf, bitmapped graphics files
\usepackage{epsfig} % for postscript graphics files
\usepackage{amssymb}  % assumes amsmath package
\usepackage{wrapfig}
\usepackage{authblk}
\title{Robust 2D Assembly Sequencing via Geometric Planning \\ with Learned Scores}

\author{Tzvika Geft$^{1}$, Aviv Tamar$^{2}$, Ken Goldberg$^{3}$ and Dan Halperin$^{1}$%
\thanks{*Work by D.H and T.G. has been supported in part by the Israel Science Foundation
(grant no.~825/15), by the Blavatnik Computer Science Research Fund,
and by grants from Yandex and from Facebook.
A.T. was partly supported by Siemens.
K.G. is supported in part by donations from Siemens, Google, 
Toyota Research Institute, Autodesk, ABB, Knapp, Honda, Intel, 
Hewlett-Packard.
}
\thanks{$^{1}$Blavatnik School of Computer Science, Tel-Aviv University, Israel}%
\thanks{$^{2}$Technion, Haifa, 3200003, Israel. Part of the work was done at University of California, Berkeley.
}%
\thanks{$^{3}$Dept. of EECS, University of California, Berkeley}%
}

\begin{document}

\maketitle
\def\frechet{Fr\'echet\xspace}

\newcommand{\cupdot}{\mathbin{\mathaccent\cdot\cup}}

%Comments

\definecolor{gray}{rgb}{0.35,0.35,0.35}
\definecolor{blue}{rgb}{0,0,1}
\definecolor{red}{rgb}{1,0,0}
\definecolor{orange}{rgb}{0.75, 0.4, 0}
\definecolor{green}{rgb}{0.0, 0.5, 0.0}
\newcommand{\aviv}[1]{{\color{green}\textbf{Aviv: }\sf#1}}
\newcommand{\tzvika}[1]{{\color{blue}\textbf{Tzvika: }\sf#1}}
\newcommand{\ken}[1]{{\color{orange}\textbf{Ken: }\sf#1}}
\newcommand{\danny}[1]{{\color{red}\textbf{Danny: }\sf#1}}
\newcommand{\michal}[1]{{\color{gray}\textbf{Michal: }\sf#1}}

%%%%%%%%%%%
\newtheorem*{proposition*}{proposition}
\newcommand\ndbg{$N$}
\newcommand\andorg{$G$}
\newcommand\rscore{$r$}
\newcommand\mixedSquare{\text{Mixed Square }}
\newcommand\rect{\text{Rectilinear }}
\newcommand*{\RobustnessFiguresDir}{figures}%

%%%%%%%%%

%unlabeled PSPACE-hardness paper
\newcommand{\mtm}{\emph{multi-to-multi}\xspace}
\newcommand{\mts}{\emph{multi-to-single}\xspace}
\newcommand{\sts}{\emph{multi-to-single-restricted}\xspace}
\newcommand{\dtd}{\emph{single-to-single}\xspace}

\newcommand{\cte}{\emph{full-to-edge}\xspace}
\newcommand{\ctc}{\emph{full-to-full}\xspace}
\newcommand{\ete}{\emph{edge-to-edge}\xspace}

\newcommand{\AND}{{\sc and}\xspace}
\newcommand{\OR}{{\sc or}\xspace}

%tex tools
\newcommand{\ignore}[1]{}

%%% algorithms
\def\vor{\text{Vor}}

\def\P{\mathcal{P}} \def\C{\mathcal{C}} \def\H{\mathcal{H}}
\def\F{\mathcal{F}} \def\U{\mathcal{U}} \def\L{\mathcal{L}}
\def\O{\mathcal{O}} \def\I{\mathcal{I}} \def\S{\mathcal{S}}
\def\G{\mathcal{G}} \def\Q{\mathcal{Q}} \def\I{\mathcal{I}}
\def\T{\mathcal{T}} \def\L{\mathcal{L}} \def\N{\mathcal{N}}
\def\V{\mathcal{V}} \def\B{\mathcal{B}} \def\D{\mathcal{D}}
\def\W{\mathcal{W}} \def\R{\mathcal{R}} \def\M{\mathcal{M}}
\def\X{\mathcal{X}} \def\A{\mathcal{A}} \def\Y{\mathcal{Y}}
\def\L{\mathcal{L}}

\def\dS{\mathbb{S}} \def\dT{\mathbb{T}} \def\dC{\mathbb{C}}
\def\dG{\mathbb{G}} \def\dD{\mathbb{D}} \def\dV{\mathbb{V}}
\def\dH{\mathbb{H}} \def\dN{\mathbb{N}} \def\dE{\mathbb{E}}
\def\dR{\mathbb{R}} \def\dM{\mathbb{M}} \def\dm{\mathbb{m}}
\def\dB{\mathbb{B}} \def\dI{\mathbb{I}} \def\dM{\mathbb{M}}
\def\dZ{\mathbb{Z}}

\def\E{\mathbf{E}} % used for denoting expectation

\def\eps{\varepsilon}

\def\limn{\lim_{n\rightarrow \infty}}

\def\obs{\mathrm{obs}}
\newcommand{\defeq}{%
  \mathrel{\vbox{\offinterlineskip\ialign{%
    \hfil##\hfil\cr
    $\scriptscriptstyle\triangle$\cr
    %\noalign{\kern0ex}
    $=$\cr
}}}}
\def\Int{\mathrm{Int}}

\def\Reals{\mathbb{R}}
\def\Naturals{\mathbb{N}}
\renewcommand{\leq}{\leqslant}
\renewcommand{\geq}{\geqslant}
\newcommand{\compl}{\mathrm{Compl}}

\newcommand{\sig}{\text{sig}}

\newcommand{\sbs}{sampling-based\xspace}
\newcommand{\mr}{multi-robot\xspace}
\newcommand{\mpl}{motion planning\xspace}
\newcommand{\mrmp}{multi-robot motion planning\xspace}
\newcommand{\sr}{single-robot\xspace}
\newcommand{\cs}{configuration space\xspace}
\newcommand{\conf}{configuration\xspace}
\newcommand{\confs}{configurations\xspace}

% libs
\newcommand{\stl}{\textsc{Stl}\xspace}
\newcommand{\boost}{\textsc{Boost}\xspace}
\newcommand{\core}{\textsc{Core}\xspace}
\newcommand{\leda}{\textsc{Leda}\xspace}
\newcommand{\cgal}{\textsc{Cgal}\xspace}
\newcommand{\qt}{\textsc{Qt}\xspace}
\newcommand{\gmp}{\textsc{Gmp}\xspace}

% programming
\newcommand{\Cpp}{C\raise.08ex\hbox{\tt ++}\xspace}

% Generic programming code:
\def\concept#1{\textsf{\it #1}}
\def\ccode#1{{\texttt{#1}}}

\newcommand{\ch}{\mathrm{ch}}
\newcommand{\pspace}{{\sc pspace}\xspace}
\newcommand{\threesum}{{\sc 3Sum}\xspace}
\newcommand{\np}{{\sc np}\xspace}
\newcommand{\degree}{\ensuremath{^\circ}}
\newcommand{\argmin}{\operatornamewithlimits{argmin}}

\newcommand{\Gdisk}{\G^\textup{disk}}
\newcommand{\Gbt}{\G^\textup{BT}}
\newcommand{\Gsoft}{\G^\textup{soft}}
\newcommand{\Gnear}{\G^\textup{near}}
\newcommand{\Gembed}{\G^\textup{embed}}

\newcommand{\dist}{\textup{dist}}

\newcommand{\Cfree}{\C_{\textup{free}}}
\newcommand{\Cforb}{\C_{\textup{forb}}}

\newtheorem{lemma}{Lemma}
\newtheorem{theorem}{Theorem}
\newtheorem{corollary}{Corollary}
\newtheorem{claim}{Claim}
\newtheorem{proposition}{Proposition}

\theoremstyle{definition}
\newtheorem{definition}{Definition}
\newtheorem{remark}{Remark}
\theoremstyle{plain}
\newtheorem{observation}{Observation}

\def\len{c_\ell}
\def\bot{c_b}

\def\lenopt{\len^*}
\def\botopt{\bot^*}

\def\Im{\textup{Im}}

\def\rfunc{\left(\frac{\log n}{n}\right)^{1/d}}
\def\rfuncs{\left(\frac{\log n}{n}\right)^{1/d}}
\def\cfunc{\sqrt{\frac{\log n}{\log\log n}}}
\def\rtrs{\gamma\rfunc}
\def\ctrs{2\cfunc}
\def\aconn{\A_\textup{conn}}
\def\abd{\A_\textup{str}}
\def\aspan{\A_\textup{span}}
\def\aopt{\A_\textup{opt}}
\def\ao{\A_\textup{ao}}
\def\acfo{\A_\textup{acfo}}
\def\binomial{\textup{Binomial}}
\def\twin{\textup{twin}}

\def\aas{a.a.s.\xspace}
\def\0{\bm{0}}

\def\distU#1{\|#1\|_{\G_n}^U}
\def\distW#1{\|#1\|_{\G_n}^W}

\def\tooth{\scalerel*{\includegraphics{./../fig/tooth}}{b}}

\makeatletter
\def\thmhead@plain#1#2#3{%
  \thmname{#1}\thmnumber{\@ifnotempty{#1}{ }\@upn{#2}}%
  \thmnote{ {\the\thm@notefont#3}}}
\let\thmhead\thmhead@plain
\makeatother

\def\todo#1{\textcolor{blue}{\textbf{TODO:} #1}}
\def\new#1{\textcolor{magenta}{#1}}
\def\old#1{\textcolor{red}{#1}}

\def\removed#1{\textcolor{green}{#1}}

\def\dx{\,\mathrm{d}x}
\def\dy{\,\mathrm{d}y}
\def\drho{\,\mathrm{d}\rho}

% algorithms
\newcommand{\prm}{{\tt PRM}\xspace}
\newcommand{\prmstar}{{\tt PRM}$^*$\xspace}
\newcommand{\rrt}{{\tt RRT}\xspace}
\newcommand{\rrtstar}{{\tt RRT}$^*$\xspace}
\newcommand{\rrg}{{\tt RRG}\xspace}
\newcommand{\btt}{{\tt BTT}\xspace}
\newcommand{\fmt}{{\tt FMT}$^*$\xspace}
\newcommand{\mstar}{{\tt M}$^*$\xspace}
\newcommand{\drrtstar}{{\tt dRRT}$^*$\xspace}

%\def\removed#1{}
%%% Local Variables:
%%% mode: plain-tex
%%% TeX-master: "main"
%%% End:

\begin{abstract}
To compute robust 2D assembly plans, we present an approach that combines geometric planning with a deep neural network. We train the network using the Box2D physics simulator with added stochastic noise to yield robustness scores~--~the success probabilities of planned assembly motions. As running a simulation for every assembly motion is impractical, we train a convolutional neural network to map assembly operations, given as an image pair of the subassemblies before and after they are mated, to a robustness score. The neural network prediction is used within a planner to quickly prune out motions that are not robust. We demonstrate this approach on two-handed planar assemblies, where the motions are one-step translations. Results suggest that the neural network can learn robustness to plan robust sequences an order of magnitude faster than physics simulation.
\end{abstract}

\section{Introduction}
Given a set of parts and their relative positions in a product, the assembly sequencing problem is to find a sequence of collision-free motions that will merge the separated parts into the final assembly \cite{halperin2000general}.
While the problem is PSPACE-hard in general~\cite{pspace}, efficient algorithms have been developed for restricted types of motions, which guarantee to find a valid assembly sequence when one exists. However, even under such restrictions, there may be exponentially many valid sequences and so the computational hardness remains for the task of choosing an optimal sequence (based on a desired measure). We focus on such a task in this paper -- finding a \textit{robust} assembly sequence, which we define as one having a high probability to succeed under small deviations from the planned motions. Such deviations arise in real scenarios where noise is introduced due to uncertainties in control and sensing.

% Leftover from abstract
%and the difficulty of computing motion plans for tight-fitting object insertions. However, recent developments in machine learning as well as in robust geometric computing suggest new lines of attack on the problem.

% benefits of robust sequences?
%Solving this problem is important for designing products that are easy to manufacture, and for planning of the layout of manufacturing facilities.

\begin{figure}[H]
\vspace{-5pt}
\centering
\includegraphics[width=0.45\textwidth]{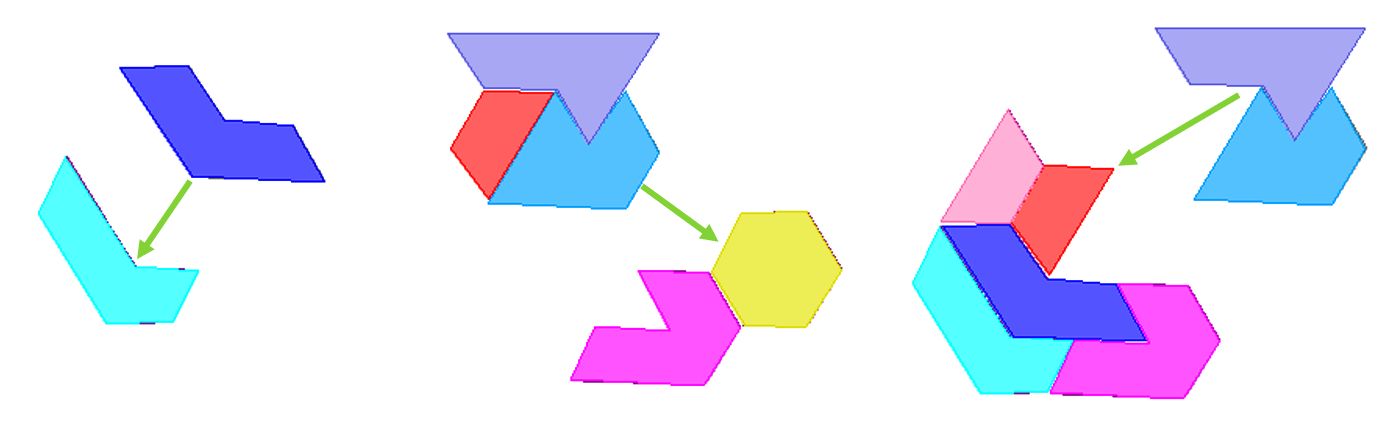}
\caption{Examples of 3 highly-robust operations. In each operation one subassembly moves as a rigid body in the direction shown while the other one is fixed.}
\label{fig:intro-robust}
\vspace{-5pt}
\end{figure}

To provide some intuition about the problem and the effects of deviations, we present examples of assembly operations and their robustness levels, starting with highly robust ones in Figure \ref{fig:intro-robust}. Note that the poly-lines along which the subassemblies in this figure are mated are characterized by wide ``funnels'', where mechanical compliance can guide subassemblies into alignment. Together with an appropriate direction for mating, these motions guide the moving subassemblies into the correct final position. We contrast these operations to ones that are not robust in the next figure:

\begin{figure}[H]
\vspace{-5pt}
\centering
\includegraphics[width=0.45\textwidth]{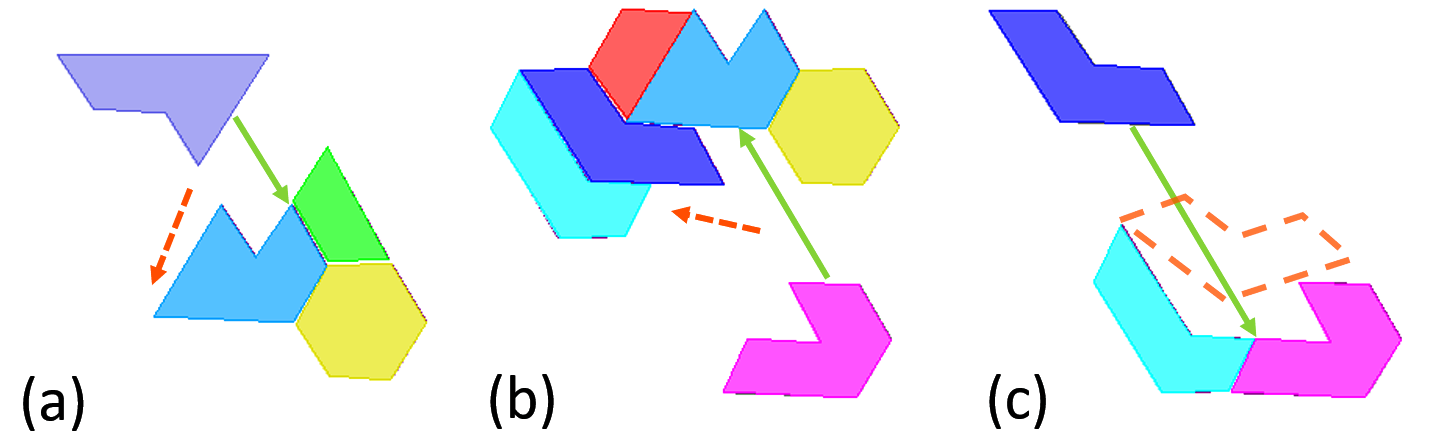}
\caption{Examples of operations that are not robust, in which noise along the planned path (shown by the green arrow) can result in undesirable motions (represented by red dashed arrows). We assign operations (a) and (b) a score of 0.5 (out of 1), and operation (c) a score of only 0.14 (see text for how it is obtained). We provide an explanation for the low scores: In operation (a) any deviation of the moving part to the left would result in it sliding away from the target along the part that is below it. The mating direction aggravates the situation (a better one would bring the moving part from the North-Northeast). In operation (b), only a single direction can be used and a similar scenario can occur if the moving part deviates left. In both (a) and (b), a deviation to the right should still allow successful completion. In contrast, operation (c) has a particularly low success rate since a deviation either way would likely result in failure: Missing right would cause the moving part to rotate and settle in a configuration (shown by the red dashed line) that can only be recovered from by fully backing up. A deviation to the left could end up the same way, since recovering from such a deviation would entail some momentum to the right.}
\label{fig:intro-not-robust}
\vspace{-5pt}
\end{figure}

\begin{wrapfigure}{r}{2.7cm}
\vspace{-15pt}
\includegraphics[width=2.65cm]{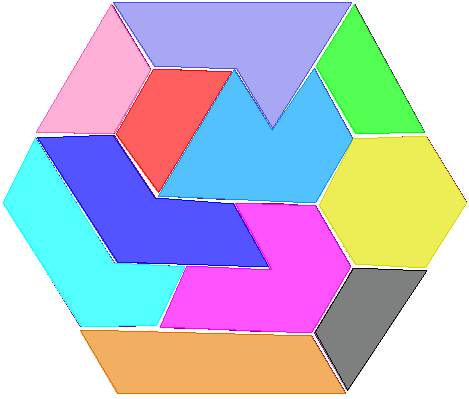}
\caption{Example of a final assembly.}
\label{fig:hex}
\vspace{-15pt}
\end{wrapfigure} 
These are just a few examples out of the exponentially many possible operations for the assembly in Figure \ref{fig:hex}. Our task is to find assembly sequences with robust operations (like those in Figure~\ref{fig:intro-robust}) by avoiding difficult configurations of parts already assembled (like those in Figure~\ref{fig:intro-not-robust}) and choosing appropriate mating directions.

In order to evaluate the robustness of assembly operations, we use a physics simulator, as we can exploit its immediate ability to capture the dynamics of the task, compared to analytic approaches (which would require handling potentially challenging intricacies, as we exemplify in Section~\ref{sec-clearance}).
% Can add citations here (as previous done....)
In particular, in this work we evaluate robustness by simulating object mating using a proportional feedback controller~\cite{franklin1994feedback} that moves an object along the assembly trajectory in the presence of actuation noise.
In this case, robustness corresponds to the ability of the planned motion to tolerate noise.
%We hypothesize that this form of robustness will manifest itself in real scenarios as well when performing similar operations. In this work we consider the simulator as the ``ground truth'' scoring function and use it to score training data and to evaluate the sequences.

Since the noise we add is random, multiple simulations are required to get the score, rendering it computationally burdensome. The effect on run time is compounded by the fact that finding a good sequence involves scoring a myriad of candidate operations. Although planning robust assembly sequences can be performed offline for a 
given assembly, efficient algorithms can facilitate efficient design for 
assembly (DfA), where small changes in geometry can increase robustness 
and planning is executed inside an optimization cycle. We therefore introduce a convolutional neural network (CNN) that has potential to efficiently identify high scored assembly operations, given their visual representation, using binary classification. 
%T.G. "that has potential to efficiently identify" - should sound more encouraging

We plan assembly sequences by obtaining a set of possible operations at each step using geometric planning and then greedily selecting the most robust one among them using the CNN. In case the CNN classifies all available operations as \textit{not} highly robust, we fall back on the physics simulator to choose the best one (though this rarely occurred in our experiments).

While our proposed method handles planar assemblies, where the motions are one-step translations (like those in Figures~\ref{fig:intro-robust}~and~\ref{fig:intro-not-robust}), it is an initial step towards handling spatial assemblies. The planar case already has a high combinatorial complexity and raises interesting challenges, as we illustrate in Section \ref{sec:hardness}.

% Say something about possibility of replacing optimization and planning components. Also maybe use robustness evaluation independently in existing systems.

\textbf{Contribution.} We present the Robust Assembly Planning (RAP) algorithm, which takes a planar assembly and returns a robust two-handed assembly sequence using one-step translations (or determines that no feasible sequence exists under such motions). RAP also considers assembly operations that are not linear (i.e., ones which bring together more than one part into an existing subassembly).

%The remainder of this paper is organized as follows: Section~\ref{sec:related} presents related work and is followed by preliminaries for assembly planning in Section~\ref{sec:prelim}. Section~\ref{sec:problem_statement} gives the problem statement, after which we illustrate the hardness of assembly sequence optimization in Section~\ref{sec:hardness}. Section~\ref{sec:the-algorithm} describes the overall RAP algorithm, including a discussion on an alternative robustness scoring methods. Sections~\ref{sec-sim} and~\ref{sec-NN} are respectively dedicated to the physics simulation and neural network.
\section{Related Work}\label{sec:related}
%\noindent\textbf{Assembly Planning and Optimization}
\subsection{Assembly Planning and Optimization}
Assembly planning is a well-studied problem in manufacturing and robotics. Some early planners employ a potentially exponential running-time generate-and-test approach, which enumerates all possible operations and tests their feasibility~\cite{gen-and-test} while others pose questions to a human expert in order to establish precedence between operations~\cite{user-queries}. A pioneering work by Wilson and Latombe~\cite{wilson1994geometric}, on which our geometric planner is based, introduces the \textit{non-directional blocking graph}, which uses geometric reasoning and examines assembly operations in the space of the allowable assembly motions. This approach avoids the inherent combinatorial trap and leads to polynomial time algorithms for motions such as one-step translations and infinitesimal rigid motions~\cite{halperin2000general}.

Given the algorithmic success in finding feasible assembly sequences, a natural goal is to find ones that meet desired properties or optimality criteria. Goldwasser et al.~\cite{goldwasser1996complexity, goldwasser-thesis} show that optimizing assembly sequences under simple motions can be NP-hard even to approximate (we list some of their cost measures in Section~\ref{sec:hardness}). Assembly optimization is therefore typically solved with heuristic techniques. As the subject is vast, with many possible optimization criteria~\cite{constraints-survey}, we refer to the following surveys:~\cite{jimenez2013survey, ghandi2015review}.
%Can also add something about sampling based, for example that it doesn't give a way to systematically enumerate options (this is said in some paper)

\subsection{Robustness}
A great deal of work revolves around planning and executing fine motions to handle challenging peg-in-hole style assembly tasks (e.g~\cite{kim2012hole-tight-task, tang2016autonomous-tight-task}). We, on the other hand, seek to plan inherently robust sequences that minimize difficult tasks during assembly. We mention a few works that plan sequences using quality measures reminiscent of ours. Heger~\cite{heger2008generating} focuses on the assembly environment by considering planar assemblies inside a constrained workspace, where robots would be operating. The work combines symbolic and motion planning to generate sequences and greedily optimizes them for two relevant measures: clearance along the path around the part being inserted and the reachability of its docking locations in each step. Moving the partial assembly between steps is allowed for improving these measures. Wan et al.~\cite{wan2018assembly} optimize sequences based on a score for the ease of 3D part insertion derived from contact normals (ignoring potential disturbances that can occur along the motion). Anders et al.~\cite{anders2018reliably} also learn from simulation to plan pushing motions for planar objects. They approach uncertainty with conformant planning, in which actions are sequenced to guarantee a successful configuration without intermediate sensing. They acquire a belief-state transition model for pushing squares into an arrangement by training a random-forest regressor.

\begin{comment}
As for robustness, 
Previous work relevant to assembly planning falls into two
main categories: approaches that tackle the problem as a
sequencing problem on an abstract symbolic level, and ones
that consider fine-grained motions of the robots involved
\end{comment}

\subsection{Learning}
The use of neural networks in assembly planning has been explored in~\cite{hong1995neural,sinanouglu2005assembly,chen2008three}. In these works, supervised learning was used to learn a mapping from geometric features in an assembly to a complete assembly sequence, with the goal of replacing the search computation with a learned heuristic. In our work we use neural networks differently: we use them to replace physics simulation in predicting the robustness of an assembly operation, from a visual representation of the proposed assembly. We incorporate the predictions from this neural network into the assembly planning framework. 

Neural networks have been used to learn controllers for assembly tasks, using inverse models~\cite{kim2000neural}, and reinforcement learning~\cite{thomas2018learning}. Neural networks have also been used for solving various planning problems in robotics, such as motion planning~\cite{pfeiffer2016perception,ichter2017learning}, grasping~\cite{mahler2019learning}, and pose estimation~\cite{xiang2017posecnn}. In this work, we apply deep CNNs to learning the robustness of an assembly operation, namely the motion that puts two subassemblies together to form one larger subassembly.

\begin{comment}{
From motion space approach:
Recently it has been shown that several variants of cost measure optimization in
assembly planning are hard, even in their approximate versions [13], [15]. We cite here
one hardness result from [14], for a problem that at first glance seems very simple.}
\end{comment}

\section{Preliminaries}\label{sec:prelim}
\subsection{Assembly Planning}
An \textbf{assembly} is a collection of bodies (also called parts) in some given relative placements, such that no two bodies overlap. A \textbf{subassembly} is a subset of the bodies composing an assembly $A$ in their relative placements in $A$. We say that two subassemblies are \textbf{separated} if they are arbitrarily far apart from one another. Given an assembly $A$, an assembly operation is a motion that merges $s$
($s \geq 2$) pairwise separated subassemblies of $A$ into a new subassembly of $A$. During this motion, each subassembly moves as a single rigid body and no overlapping between bodies is allowed.~\footnote{As is common in \emph{assembly planning} we allow motion of parts in contact, such as one part sliding over the other, but forbid overlap of the interior of parts during the motion.} In this paper the assembly operations are \textbf{two-handed}, i.e., every operation merges exactly two subassemblies. The reverse of an assembly operation is a \textbf{partitioning} operation. We denote both types of operation as a tuple $(S_1, S_2, m)$, where $S_1, S_2 \subset A$ are the subassemblies merged/partitioned by the motion $m$. An \textbf{assembly sequence} is a total ordering on assembly operations that merges the separated parts composing an assembly into this assembly.

A common approach in assembly planning is \textbf{assembly-by-disassembly}, whereby a \textit{dis}assembly sequence is obtained and then reversed. As the final assembly configuration is the most constrained, the approach reduces the search space and naturally offers valid operations. Disassembly consists of partitioning, first the input assembly $A$ into subassemblies and then, recursively, the resulting subassemblies that are not individual parts. When only considering geometric feasibility, and assuming that the parts are rigid, the sequences are symmetric, though this is not true if we also care about robustness (e.g., the peg-in-hole scenario). Nevertheless, the approach is suitable for our setting, since we treat partitioning operations as the respective assembly operations when evaluating robustness. Due to this interchangeability of assembly and disassembly, we sometimes simply use the term \textbf{sequence}.

% Maybe add a statement that no backtracking is required
\subsection{The Motion Space Approach}\label{sec:ndbg}
\begin{figure}[H]
\vspace{-5pt}
\centering
\includegraphics[width=0.335\textwidth]{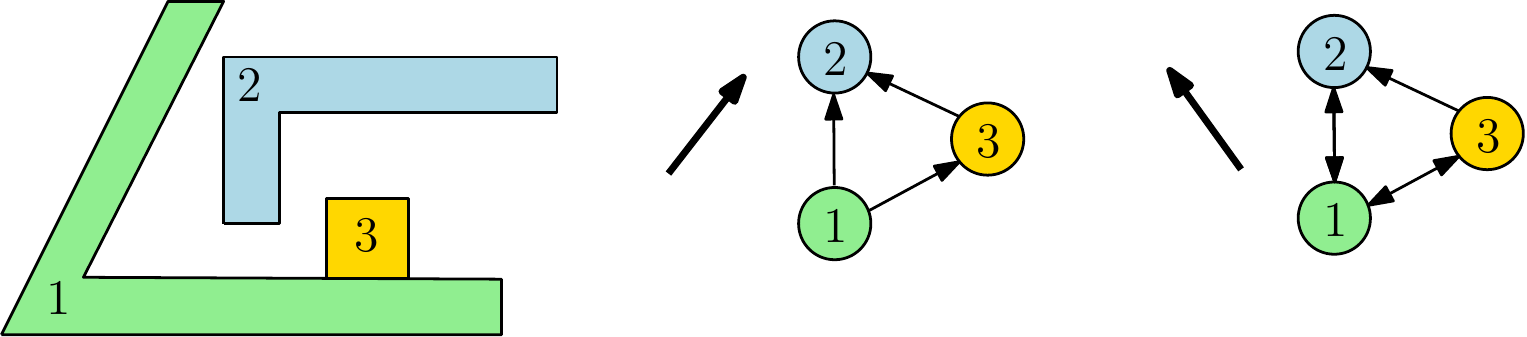}
\caption{DBGs for two directions used for one-step translation~\cite{halperin2000general}.}
\label{fig:dbgs}
\vspace{-6pt}
\end{figure}

To complete the description of the disassembly process, we give an overview of the partitioning procedure using the so-called \textbf{motion space approach}~\cite{halperin2000general}. The \textbf{motion space} is defined to be the space of parametric representations of all allowable motions for partitioning operations: every point in it uniquely defines a path of the subassemblies moved by an operation. A key concept in this approach is the \textbf{directional blocking graph} (DBG)~\cite{wilson1994geometric}. Given a specific motion, a DBG is a directed graph that represents the blocking relation between parts: each node represents a part and an edge from part $A$ to part $B$ exists if applying the motion on part $A$ results in a collision with part $B$ (see Figure~\ref{fig:dbgs}). Given a DBG, it is sufficient to compute its strongly connected components in order to find a valid partition that uses this motion (or determine that no such partition exists, as is the case when the DBG is strongly connected).

An important insight is that a single DBG can represent many motions, as the blocking relations are not necessarily affected by small changes to a given motion. This fact allows for decomposing the motion space into regions called \textbf{cells}, such that all motions in the same cell induce a single DBG. The decomposed motion space is called the \textbf{non-directional blocking graph} (NDBG)~\cite{wilson1994geometric}. Once an NDBG is obtained for a given assembly, partitioning it involves finding a feasible partition for any of the DBGs (i.e., motions) it contains, meaning the NDBG captures all the geometric information required for partitioning. Its structure, particularly the number of cells in the decomposition, is therefore a critical run-time factor that depends on the type of allowed motions. For one-step translations (in both 2D and 3D) this number is polynomial in the number of parts in the assembly (and their complexity), which allows for finding an arbitrary assembly sequence in polynomial time (exact bounds for our setting appear in Section~\ref{sec-complexity}). For more details on NDBG construction and the application of the motion space approach for a few families of motions see~\cite{halperin2000general}. %\tzvika{To Danny: I think the last two paragraphs are fine overall, though I'm unsure of whether there are too many details on the DBG.}
%intro found in "motion space approach" might be more suitable here
%As we will see, searching for a good sequence in this way is not quite so simple

\section{Problem Statement} \label{sec:problem_statement}
Given a planar assembly with $n$ parts, we would like to find the most robust two-handed sequence for it, where the motions are one-step translations, such as those in Figure~\ref{fig:intro-robust}. To evaluate a sequence for robustness we use the simulator to score the individual assembly operations in it, resulting in the scores $r_1, \ldots, r_{n-1} \in [0,1]$ (note that a two handed sequence for $n$ parts always has $n-1$ steps). As we define the robustness of a single operation as its success probability under noise, the score we assign to the whole sequence is the product of these individual scores, $R \vcentcolon= \prod r_i$, and our goal is to find a sequence that maximizes it.

As the search space involved here is potentially exponential, we restrict it to a polynomial one by greedily choosing assembly operations and in a few other ways, which will be described in the sequel.
%First, we only consider one partition for each NDBG cell, and in particular we only consider one direction of motion out of a possible multitude of %directions. Having one representative from each cell, we score each of these partitions, and contribue with the single bst scorte.

\section{The Hardness of Finding Optimal Sequences} \label{sec:hardness}
In this section we illustrate the hardness of finding optimal sequences by considering a generic cost function, which takes an assembly sequence and returns a real value as its cost.
% Define function
\begin{proposition*}
Finding an optimal assembly sequence is NP-hard.
\end{proposition*}
\begin{proof}
We give a reduction from the PARTITION problem, which is: given $n$ positive integers $a_1, \ldots, a_n$, decide whether they can be partitioned into two subsets whose sums are equal. Given such an instance $I$, we define an assembly $A$ to be a row of $n$ axis-aligned rectangles with the same height, where the width of rectangle $i$ is $a_i$. We define a cost function that only takes into account the last operation, which merges two subassemblies, $S \subset A$ and $A \setminus S$, into the final assembly: The cost is the absolute difference between the sum of the widths of the rectangles in $S$ and the corresponding sum for the rectangles in $A \setminus S$. Clearly, under this cost function $I \in$~PARTITION if and only if the optimal cost of assembling $A$ is 0.

%$|\sum_{r \in S} \textnormal{width of }r - \sum_{r \in A\setminus S }\textnormal{width of } r|$.
\end{proof}

A possible motivation for the cost function used is a desire to have the weights of the two subassemblies be close for balance reasons. Note that PARTITION was also used to show the NP-hardness of finding a feasible sequence for a planar rectilinear block puzzle, where each block is allowed to move multiple times using translations that are also rectilinear~\cite{chazelle1984complexity}.

More NP-hard (dis)assembly optimization goals under one-step translations include minimizing the number of directions used to mate subassemblies and minimizing the number of parts that need to be removed from an assembly in order to remove a key part~\cite{goldwasser-thesis}.
These hardness results motivate us to follow an approximate solution to robust assembly sequencing, which we describe next. 

% Other options to include here is minimizing re orientations or direction-based cost that I proved

\section{The RAP Algorithm}\label{sec:the-algorithm}
We first give an overview and then explain how the planning, physics simulation, and CNN are used. Given a planar assembly $A$, we wish to find an assembly sequence for it with the best robustness score. Due to the hardness of optimizing sequences we use a greedy approach, which selects the most robust assembly operation available at each step. To this end, we first construct the NDBG for $A$ and then apply Algorithm~\ref{alg:RAP} on the complete assembly. The algorithm recurses on subassemblies introduced by partitions, following the assembly-by-disassembly approach.

%Try this:
%https://tex.stackexchange.com/questions/145760/how-may-i-remove-endif-and-endfor-in-algorithmicx
\vspace{-5pt}
\begin{algorithm}
    \SetKwInOut{Input}{Input}
    \SetKwInOut{Output}{Output}

    \Input{Subassembly $S \subseteq A$}
    \Output{Assembly sequence for $S$}
    P $\gets$ List of feasible partitions for $S$ obtained from NDBG \label{alg1:ndbg}
    
    \If {no feasible partitions exist}
      {
        return failure
      }
    $(S_1, S_2, \overrightarrow{d}) \gets$ \textbf{select\_best\_partition}(P) \label{alg1:scoring}
    
    Output $(S_1, S_2, \overrightarrow{d})$
    
    Continue recursively on $S_1$ and $S_2$ (when more than a single part remains)
    
    \caption{A generic greedy approach for finding a robust assembly sequence.}
\label{alg:RAP}
\end{algorithm}
\vspace{-5pt}

%We proceed to address the algorithm's components as follows:
Section~\ref{sec-NDBG-query} explains how the NDBG is used (line~\ref{alg1:ndbg}). As for the procedure \textbf{select\_best\_partition} (line~\ref{alg1:scoring}), we have two implementations: an efficient CNN-based one, which we use for RAP, and baseline that only uses the physics simulator to score partitions (returning the one with highest score). We describe the former in Section~\ref{sec-scoring}, providing more details on the simulator in Section~\ref{sec-sim}, and compare the two implementations with experiments in Section~\ref{sec-exp}.
%Note that in general the procedure can be modified to address other quality measures than robustness.

\begin{comment}
Given a planar assembly, the algorithm first constructs an NDBG for it and then continuously uses it to grow an AND/OR graph representing various disassembly sequences, following the assembly-by-disassembly approach. Assembly operations that arise in the sequences are scored in order to find the most robust sequence. \tzvika{Maybe it should be made clear how scores are saved in the AND/OR tree in the preliminaries}
\end{comment}

\subsection{The Use of the NDBG} \label{sec-NDBG-query}
% Structure of NDBG - This section seems to be the best place to put it, though in intro to the overall approach and the preliminaries can also work.

%\michal{Add an example referring to figure of what we get from the NDBG}
%\michal{Explain why we need G and N, or one sentence on why we need it}
%Given a subassembly $S \subseteq A$, we would like to obtain a list of possible assembly operations, as a list of partitions, each having the form $(S_1, S_2, \overrightarrow{d})$
Since the motions we allow are one-step translations, a single angular parameter representing the direction of the translation defines the motion. Our NDBG is therefore a unit circle ${\cal S}^1$ that is decomposed into an arrangement of vertices and arcs. To construct it we use the implementation of Fogel and Halperin~\cite{fogel2013polyhedral} (adapted to 2D), which is performed once at the beginning of the algorithm.

To obtain a choice of possible partitions for a given subassembly $S \subseteq A$, we examine each cell in the NDBG and its associated DBG (restricted to $S$). As outlined in Section~\ref{sec:ndbg}, each DBG encodes all possible partitions for the directions of motion represented by the cell. In order to keep the overall running time polynomial,  we consider only a single partition per DBG (out of a potentially exponential number, e.g., when no part is blocked by another part in some direction). This entails choosing two subassemblies, $S_1, S_2 \subset S$, which is done arbitrarily among the valid options, and a direction, for which we have the freedom of choice only when the cell is an arc of ${\cal S}^1$. Such a cell represents a range of directions along which $S_1$ and $S_2$ can be mated and we select the middle of that range. This is a heuristic, which in many cases would result in a large clearance for the assembly operation, relative to other directions in the range.\footnote{For this to hold, adjacent NDBG cells that admit the same subassemblies in their partitions need to be merged. Indeed this is what our algorithm does. We omit the details.} While experiments mostly support this choice of direction, a more sophisticated selection can be considered in future work (e.g. by learning it as well). In Section \ref{sec-clearance} we further discuss clearance as a robustness measure.

We make two remarks on the restriction of one partition per DBG: First, our approach remains complete, as all directions are considered. Another fact that remains is that the space of feasible sequences is still potentially exponential. What reduces it further to a polynomial one is the greedy selection of the best partition.
%Note that the NDBG for $A$ can be reused for all its subassemblies as each DBG is still valid when reduced to the remaining parts.

%can make a note about herustics being applied here, possibly domain specific
%In both cases there is room for more sophisticated selection, which we leave to future work.

%Note that if no feasible partition is found after trying all the cells, we stop and report the assembly to be \textit{interlocked}, i.e., an assembly sequence using one-step translation does not exist.

% Can discuss the following details which are a bit TMI: The need to subgraph the DBG and the fact that we test all of them, instead of finding just one partition  (possible use the sentence "... it can be queried to generate one assembly sequence, or several, or all..." from an earlier version.

%Maybe use procedure like pg 7 of motion space approach
%1) Subgraph
%2) Select partition
%3) Direction - if arc then middle direction should provide more clearance in most cases

%motion of one SA vs both issue
\subsection{Scoring and Selecting Partitions} \label{sec-scoring}
%The robustness score assigned to an assembly operation is a real value $\rscore{} \in [0, 1]$ representing the probability that the operation succeeds under small deviations from the planned motion trajectory. While we can obtain this score by simulating a noisy controller that performs the operation, we introduce the CNN to make scoring more efficient and practical. 
%While other methods can be used to obtain a score that adheres to our robustness definition, we consider the score given by the simulation as the ground truth.
%We now describe how we combine them in order to choose a robust assembly operation. Full details on the simulator and the CNN are provided in Sections~\ref{sec-sim} and~\ref{sec-NN} respectively.

As a preprocessing step, the CNN is trained on examples that are scored by the simulator and learns to classify operations as either \textit{high}, i.e., having a score in the range [0.95,~1], or \textit{not high}. Given a list of assembly operations as an argument to \textbf{select\_best\_partition} in the main algorithm, we first classify them using the CNN. If there are operations classified as \textit{high}, we choose the one with the highest \emph{confidence}, as measured by the softmax output of the network~\cite{geifman2017selective}.
% \danny{define/explain confidence and/or give a reference}. 
Otherwise, we resort to the simulator, scoring all partitions with it, and choose the one with the highest score. While a finer output from the CNN might be more desirable, we find that our use of binary classification is suitable, as the simulator does not need to be queried in almost all our experiments (see Section~\ref{sec-exp}).

\subsubsection{Clearance as an Attempted Analytic Robustness Measure}\label{sec-clearance}
\begin{figure}[H]
\vspace{-4pt}
\centering
\includegraphics[width=0.3\textwidth]{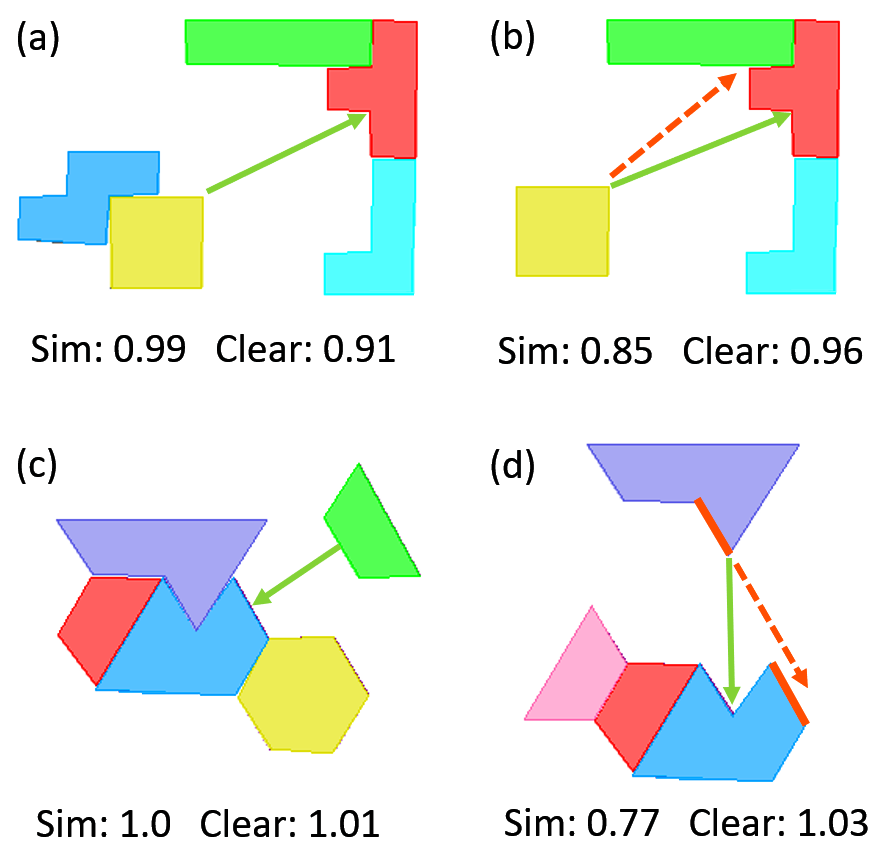}
\caption{Examples where clearance does not correlate with robustness.}
\label{fig:clearance}
\vspace{-4pt}
\end{figure}

In this section we introduce clearance, an alternative \textit{analytic} scoring method, and present its pitfalls. We define this score as the average minimum distance between the subassemblies during the mating motion.
%Maybe cite that this measure is used in motion planning
Intuitively, a higher clearance should minimize physical interaction between subassemblies that could hinder the success of the operation under noise. Indeed, we observe an overall positive correlation between this score and the simulator score, but we also find that it can mislead, as we show in Figure~\ref{fig:clearance}.
While (d) has the highest clearance score among the four examples, it has the lowest simulator score. A deviation of the moving part to the right that brings the marked edges in (d) into contact, would result in an almost unrecoverable slide away from the target. This possible failure is not captured by clearance, as (c) has a slightly lower clearance even though its shape makes it robust to similar failures. Similarly, if we had to decide between (a) and (b) based on clearance, we would also end up forgoing the robust operation for one that is not. In this case, the part above the square in (a) reduces clearance, but allows the subassembly to slide to the correct place in case it misses to the left and touches the top rectangle. The above cases exemplify just a few intricate details that would have to be accounted for in an analytic solution and therefore encourage the use of a simulator.

\subsection{Time Complexity} \label{sec-complexity}
Let $n$ be the number of parts in the input assembly $A$ and $q$ be the maximal number of edges in a single part. The NDBG has $O(n^2)$ cells and we compute it once in the beginning using $O(n^2(\log n +q^2))$ time~\cite{halperin2000general}. In line~\ref{alg1:ndbg} we obtain partitions by computing the strong components of each cell's DBG. This allows obtaining a partition using $O(n^2)$ time per cell, resulting in a total number of $O(n^2)$ partitions (one per cell) for the subassembly. In line~\ref{alg1:scoring} we score the partitions by feeding their images to the CNN and (if required) by simulating the assembly operations they represent. Let $t(n, q)$ be the time bound for scoring a single partition in this manner (a polynomial that mainly depends on the simulator's internals). We thus require $O(n^2(n^2+t(n, q)))$ time for lines~\ref{alg1:ndbg}-\ref{alg1:scoring}.
These lines are repeated exactly $n-1$ times, as each time they run an assembly operation is chosen and that is the length of the sequence. The overall run time is therefore $O(n^2q^2+n^3(n^2+t(n, q)))$.
%= O(n^2(q^2+n(n^2+t(A))))

%The described query can be done in $O(n^4)$.
%\subsection{Complexity Analysis} \label{sec-complexity}
%Either here or before give a formal definition of the input A = {P1,..}, define q and then state the time below and cite motion space approach

% NDBG construction - O(n^2(log n +q^2))
% Space requirement 2^n SAs, for each we have have n^2 partitions
% partitioning: n^4

\section{Physics Simulation} \label{sec-sim}
To evaluate robustness, we use the Box2D physics engine \cite{catto2011box2d} to simulate a controlled assembly motion of two objects with actuation noise. We start the simulation when the two objects are separated, and apply translational forces on one object to drive it to an assembled configuration, while the second object is held fixed. Rotational forces are also applied, though only for correcting the orientation of the moving object. We used a proportional feedback controller~\cite{franklin1994feedback} to track a linear path connecting the object's initial position and the goal (see~Figures~\ref{fig:intro-robust}~and~\ref{fig:intro-not-robust}).
%We also add actuation noise to the controller, resulting in a stochastic system.

We add actuation noise at each control loop iteration as follows: Let $F$ be the correct force that the controller should currently apply. We add to $F$ a random noise component drawn uniformly from the interval $[-\eta\left\lVert F\right\rVert, \eta\left\lVert F\right\rVert]$ on the axis perpendicular to $F$ and output the sum as the noisy force, resulting in a stochastic system. Currently we set $\eta=9$ as it gives a reasonable amount of noise when visualizing the operations while also resulting in a sufficiently varied distribution of robustness scores.

We measure the robustness of the operation as the success rate of the controller in driving the object sufficiently close to the goal within a fixed time, after performing 100 trials. At the beginning of each trial, we place the moving object such that it is completely outside of the bounding box of the static object (aligned with the direction of motion) and is also located at least a fixed distance away from it.

\section{Neural Network}  \label{sec-NN}

Using a physics simulator to predict the success probability of an assembly operation, as outlined in the previous section, adds significant computational overhead to the planning algorithm. We hypothesize, however, that in a practical setting, it is possible to exploit similarities between different assemblies to reduce the computational burden. For example, it may be that peg-in-hole type assemblies are particularly difficult for our robot, and we can identify such structures and avoid them without requiring extensive simulations. Here, we propose a general approach for identifying such structural properties by using supervised learning~\cite{Bishop2006Pattern}.

In our approach, we generate $N$ random assembly operation instances, $x_1, \dots, x_N$. A single assembly operation instance $x_i$ includes the initial, unassembled, position of two subassemblies, and their assembled position. For each instance $x_i$, we use the physics simulator to obtain a robustness score $y_i$. We propose to use supervised learning to learn a mapping $f$ from some features of the assembly instance $\phi(x)$ to their score $y$, such that during planning, we can use $f$ instead of the physics simulator to obtain robustness scores.

Selecting features that are relevant to the robustness score is not trivial. Here, we build on the recent success of deep convolutional neural networks (CNNs;\cite{goodfellow2016deep}) in automatically learning features from image input. We observe that in a 2D setting, an image of the subassemblies contains all the geometric information that is input to the physics simulator, and is therefore sufficient, in principle, for a CNN to decode features relevant to predict robustness. For each instance $x_i$, we therefore generate two images (see Figure~\ref{fig:CNN}): \textbf{start image}, where the subassemblies are in their initial position, and \textbf{goal image}, where the subassemblies are assembled. These images are input to a CNN, which predicts whether the operation has high robustness.

\subsection{CNN Architecture}
\begin{figure}[H]
\vspace{-5pt}
\centering
\includegraphics[width=0.45\textwidth]{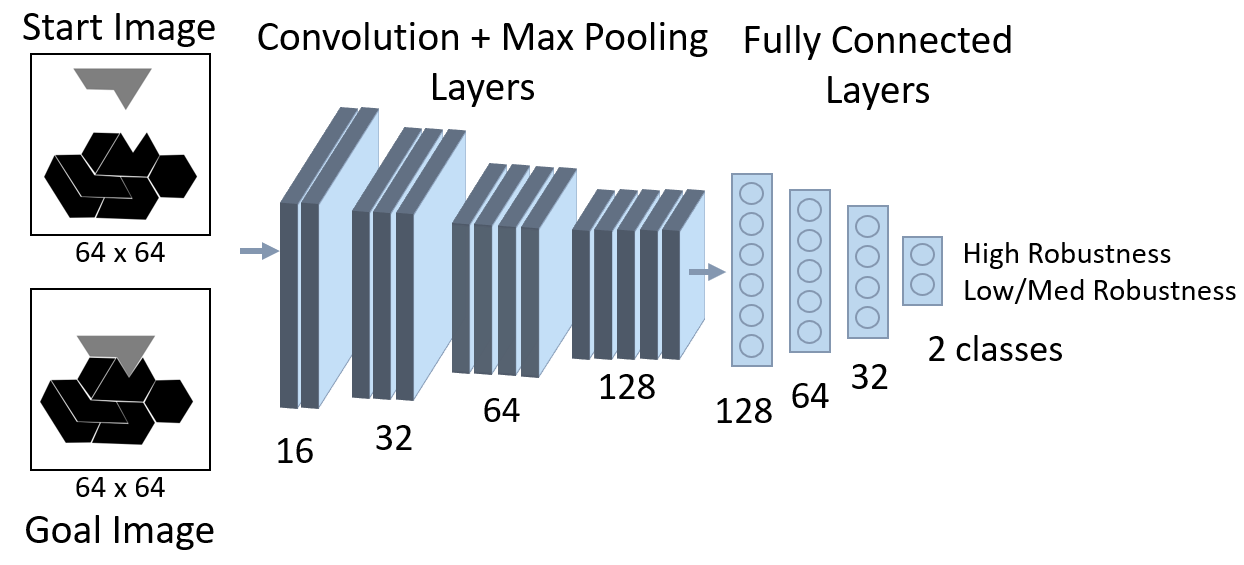}
\caption{CNN architecture for predicting robustness. Given $64 \times 64$ grayscale images of the starting position and goal position of two subassemblies, the network predicts whether the assembly operation is highly robust or not (see text for details). The CNN is composed of four convolution layers with $4\times 4$ kernels, ReLU activations and max-pooling, followed by three fully connected layers with ReLU activations, and a final linear connection to the output layer.}\label{fig:CNN}
\vspace{-5pt}
\end{figure}

CNNs have been previously used for various geometric tasks such as shape recognition~\cite{su2015multi}, pose estimation~\cite{xiang2017posecnn}, and evaluating robotic grasp success~\cite{mahler2019learning}. Here, we follow a similar approach, and design a CNN for evaluating assembly robustness, as described in Figure~\ref{fig:CNN}. Our input is represented as a 2-channel $64 \times 64$ image, where each channel corresponds to a grayscale rendering of the start and goal images. In principle, this image contains all the information about the planned trajectory and the geometry of the parts involved, and we expect that a CNN-based architecture would be able to extract this information to predict a robustness score.

We use the popular CNN architecture of several convolutions and pooling layers, followed by fully connected layers with dropout~\cite{goodfellow2016deep,lecun1998gradient,krizhevsky2012imagenet}, as depicted in Figure~\ref{fig:CNN}. For the output, we found that the numerical value of the robustness score was hard to predict accurately, and instead we quantize the score into 2 classes: \textit{high}, corresponding to the score range [0.95, 1], and \textit{not high}. We trained our networks in PyTorch~\cite{paszke2017automatic}, using Adam~\cite{kingma2014adam} with default parameters to optimize the cross-entropy loss~\cite{goodfellow2016deep}.

%We use a softmax output layer to predict the probability of each robustness class given the input images.

% Note that in our setting, the direction of movement in the assembly is important for predicting the robustness score. The CNN needs to extract this information from the two images, which in principle may be difficult if the objects are far away, due to the local computation implied by the $4 \times 4$ convolution kernels \danny{explain or give a ref}. The depth of the network, and the pooling layers, mitigate this issue, and we did not observe in practice any difficulty related to the initial distance between the subassemblies.
% \danny{expland the discussion on the initial distance}

%We also experimented with different number and sizes of the layers, but did not notice any significant change in performance.
%\tzvika{Maybe be more specific and say that varying the distance did not affect performance significantly?}

%Full details are provided in the appendix. %\tzvika{Might fill out if time permits}

\begin{figure}[H]
\vspace{-4pt}
\centering
\includegraphics[width=0.365\textwidth]{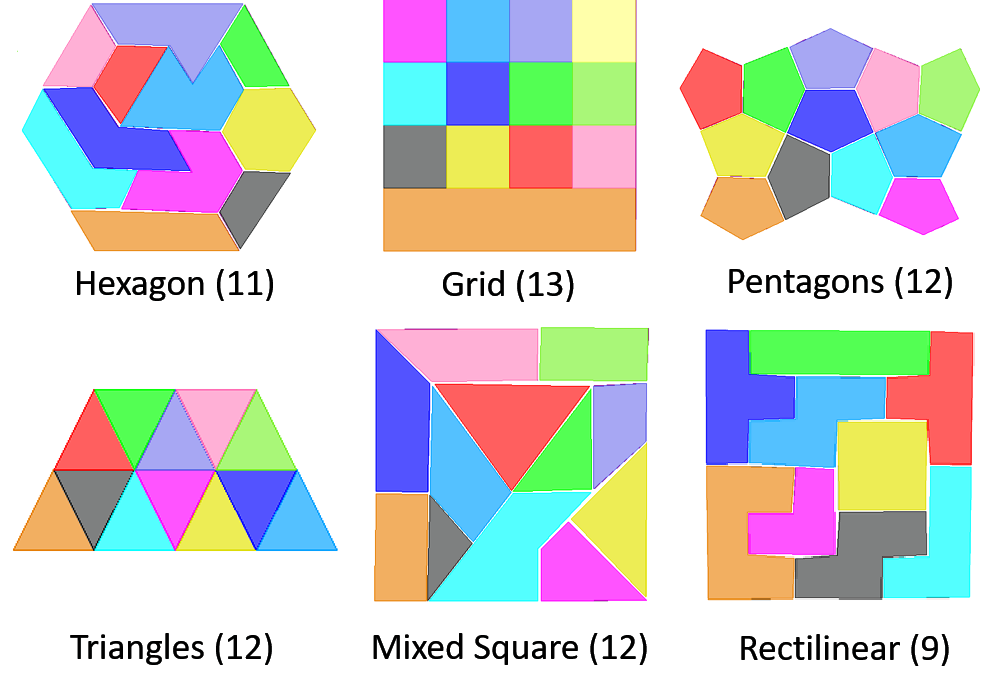}
\caption{The assembly tasks in our experiments.}
\label{fig:assemblies}
\vspace{-4pt}
\end{figure}

\subsection{Dataset Normalization}
In our experimental setting, gravity is not considered and therefore the robustness score is invariant to the orientation of the operation as a whole. We therefore preprocessed the input images by rotating them such that the moving subassembly translates downward, which resulted in a small performance gain. In the absence of such symmetry, similar gains can be obtained by increasing the data size. 
% The images that we generate for each instance are rotated such that the moving subassembly translates downward. The rationale behind this normalization is that robustness is invariant to the orientation of the operation as a whole, since interaction between the subassemblies remains the same (and gravity is not considered). A new instance encountered by the CNN whose only major difference from a previously seen instance is the orientation should then be more likely to be classified correctly. Indeed, we find that this normalization results in slight performance gains.

\section{Experiments} \label{sec-exp}
\begin{comment}
Tables:
NN accuracies

Main algorithm table:
Columns: Assemblies + average
Rows for robustness: Sim score, NN score, Optimal score, Random score
Separator row
Time comparison

(Num sim calls and work not really interesting here
Inside simulator section need to say that optimization save an average of X trials.
Can discuss rectilinear and its sim calls separately)
\end{comment}
%\subsection{Setup and Implementation Details}

We now evaluate the prediction accuracy of the CNN and RAP's run time and sequence robustness.
Figure~\ref{fig:assemblies} shows the assembly tasks in our experiments and the number of parts in each of them. We chose to analyze these puzzle-like assemblies due to their highly constrained nature, which increases the interaction between parts and requires careful planning.

\subsection{Performance of the CNN}
One question that arises when using a machine-learning approach to predicting robustness is how well the trained model generalizes to assemblies not in the training data. We investigate this question by performing 6-fold cross-validation on the data obtained from the assemblies in Figure~\ref{fig:assemblies}, consisting of 40,000 examples for each of them.\footnote{We obtain many examples for each assembly using its NDBG by exhaustively considering all possible partitions for each subassembly, as opposed to choosing just one like in RAP's online phase (i.e. we generate all possible sequences).}
We test the CNN model on data for each one of the assemblies in turn, while using the rest for training.
%Each time we test the CNN on data from one assembly while using the rest for training.
In Table~\ref{nn-table} we report the accuracy in predicting robustness. These results suggest that the CNN can indeed generalize and predict robustness for assemblies it had not seen during training. Thus, with sufficient training data obtained off-line, we can learn an effective model for predicting robustness during online planning, thereby reducing planning time.

\begin{comment}
Using the datasets described above, we evaluate our CNN model using an all-but-one approach, i.e. a different CNN is evaluated on each assembly such that it is trained on all assemblies other than the one it is evaluated on.
\end{comment}

\begin{comment}
For each assembly all assembly operations were exhaustively generated, as described in section \tzvika{6 B, ref}, resulting in the following number of examples for each of them:
\end{comment}

\begin{table}[H]
\vspace{-3pt}
\begin{tabular}{rccc}
\hline
\textbf{Test Assembly} & \multicolumn{1}{l}{\textbf{Accuracy}} & \multicolumn{1}{l}{\textbf{Low Robustness}} & \multicolumn{1}{l}{\textbf{High Robustness}} \\ \hline
Hexagon                    & 84.79                                 & 93.17                                       & 69.87                                        \\
Mixed Square           & 84.73                                 & 77.77                                       & 89.48                                        \\
Pentagons              & 74.93                                 & 94.17                                       & 72.09                                        \\
Rectilinear            & 82.64                                 & 80.70                                        & 90.99                                        \\
Grid                   & 81.70                                 & 70.98                                        & 90.35                                        \\
Triangles              & 93.13                                 & 88.77                                       & 96.41                                        \\ \hline
\end{tabular}
\caption{Overall accuracy of our CNN model in predicting robustness alongside individual class accuracy. Each row shows results for a CNN trained on examples from all assemblies other than the one it was evaluated on.}
\label{nn-table}
\vspace{-8pt}
\end{table}

\subsection{Performance of RAP}

%Our main objective is to evaluate the following:
%\begin{enumerate}
%    \item RAP is more efficient than using solely physics simulation.
%    \item It results in sequences with close to optimal robustness.
%\end{enumerate}

\subsubsection{RAP vs Simulator-Only Baseline}
For each assembly task we compare the time taken to find a sequence and its robustness score using RAP and the baseline. Regardless of the approach used, sequences are evaluated using the simulator (as it is our "ground truth"). Also note that when applying RAP on an assembly, we use the CNN that has not seen the data for that assembly in training.

%\begin{table}[H]
%\begin{tabular}{lllllll}
%\hline
%\textbf{Assembly} & \textbf{Hex.} & \textbf{Mixed Sq.} & \textbf{Pentag.} &\textbf{Rect.} & \textbf{Grid} & \textbf{Triangles} \\
%\hline
%Baseline & 19 & 22 & 20 & 23 & 20 & 25.50 \\
%RAP & 1.71 & 1.40 & 1.761 & 7 & 1.26 & 1.16 \\
%\hline
%\end{tabular}
%\caption{Run times (s) for finding a robust sequence.}
%\label{table:times}
%\end{table}

%\begin{figure}[H]
%\centering
%\includegraphics[width=0.51\textwidth]{\RobustnessFiguresDir/sim_nn_sim_comparison.png}
%\caption{Run time and robustness comparison between the simulator-only baseline %and RAP.\tzvika{The top graph can be replaced with a table and the bottom %completely omitted.}}
%\end{figure}

\begin{figure*}[!t]
%\vspace{-5pt}
\centering
\includegraphics[width=0.85\textwidth]{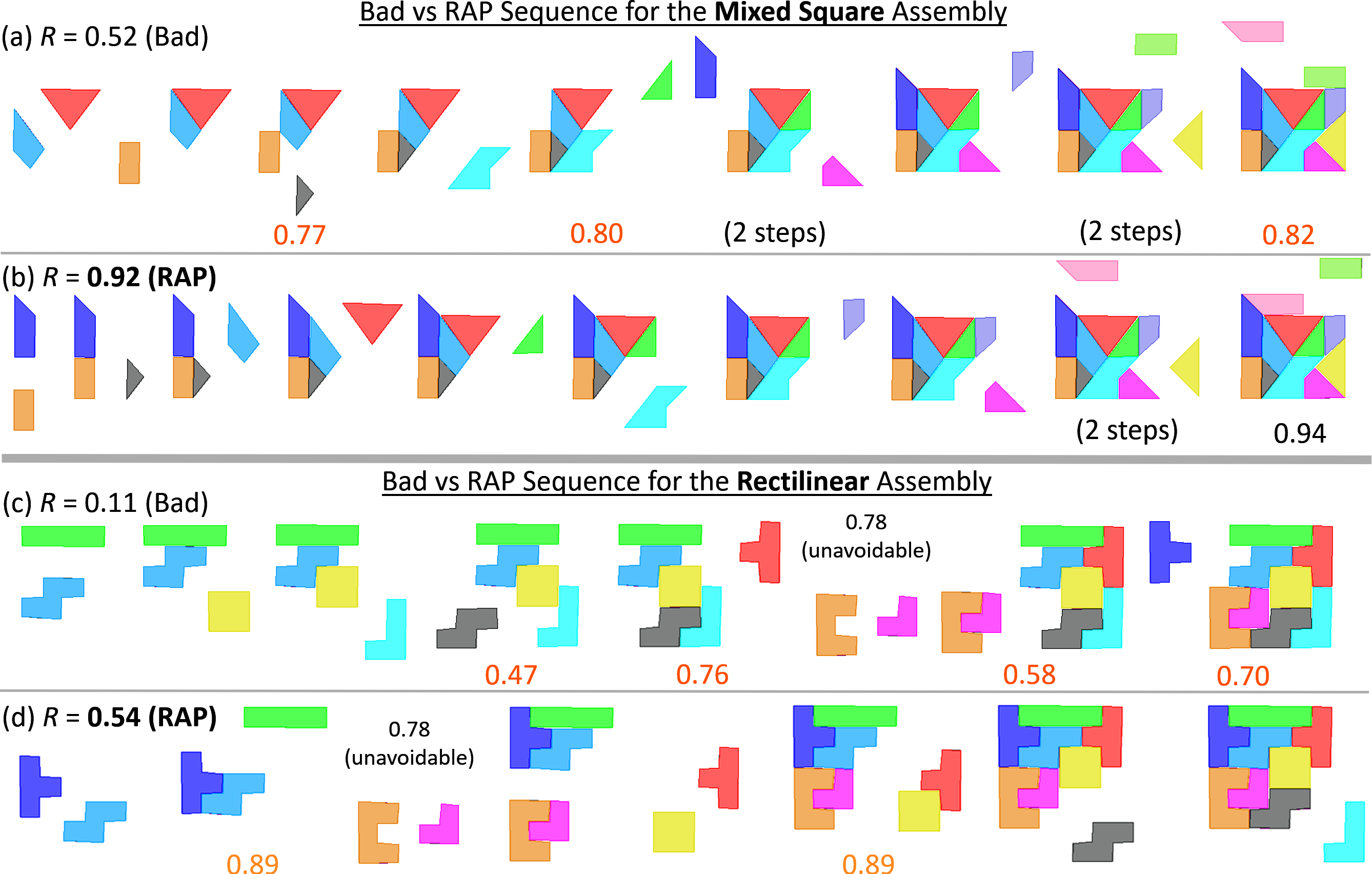}
\caption{Two sequences each for the \mixedSquare assembly, (a) and (b), and the \rect assembly, (c) and (d). For each of them the bottom sequence is returned by RAP, while the top one is an arbitrary one with low robustness. Individual motion scores are shown for non-robust steps (i.e., with a score less than 0.95). Total robustness score $R$ is the product of individual scores. Sequence (d) shows how crucial it is to consider non-linear sequences, as the operation that results in the second to last subassembly (before the S-shaped part is inserted) would be a tight insertion if we only allowed a single part to be inserted.}
%T.G. improve last sentence
%Maybe can give labels like A1, A2, B1, B2
\label{fig:seqs}
\vspace{-19pt}
\end{figure*}

The average run time on all examples for the simulator-only baseline is 21.58 seconds (with low variance) while for RAP it is 2.38 seconds. In all but the \rect assembly task RAP does not query the simulator, resulting in an average run time of 1.4 seconds for these tasks. In the \rect example, the search leads to a subassembly for which the CNN classifies all possible operations as not robust (even though one is), leading to a simulator query for them, which results in a run time of 7.0 seconds. Nevertheless, even in this case RAP improves run time by more than 3-fold (compared to 10-fold overall).

As for robustness, the scores end up very close: the average difference between the baseline and RAP is only -0.005 (scores for RAP are shown in Figure~\ref{fig:ScoreComparison}). These results indicate that the time improvements resulting from using a CNN within RAP do not come at the expense of robustness.

\subsubsection{Evaluating Robustness}
\begin{figure}[!b]
\vspace{-18pt}
\centering
\includegraphics[width=0.51\textwidth]{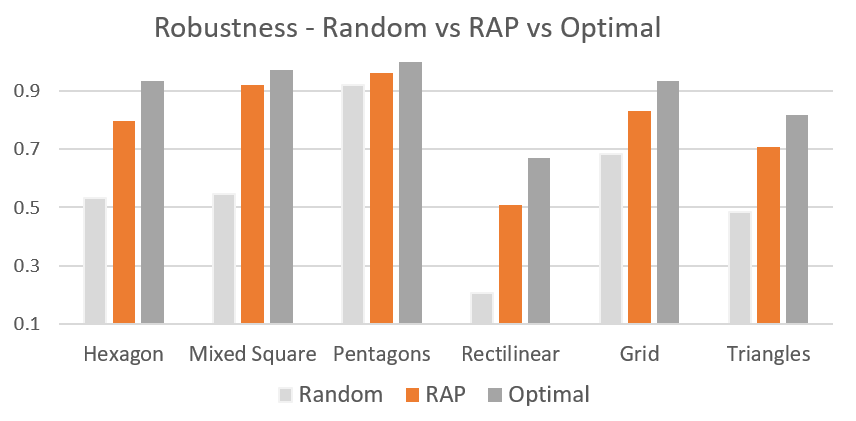}
\vspace{-15pt}
\caption{Comparison of average scores of 50 random sequences, scores of sequences obtained by RAP, and the optimal ones for the different assembly tasks. Indeed our learning-based approach offers significant robustness gains, which are close to optimal, using a fraction of the running time required by the optimal procedure.}
\label{fig:ScoreComparison}
\vspace{-1pt}
\end{figure}

 For this evaluation we must first note that absolute robustness scores do not necessarily indicate the quality of the optimization, since an assemblies' sequences can naturally tend towards a certain level of robustness. For example, there might be unavoidable steps that are not robust, such as the C-shaped part in the \rect assembly, which induces a peg-in-hole scenario (see "unavoidable" examples in (c) and (d) in Figure~\ref{fig:seqs}). To take such inherent difficulties into account, we compare the sequences obtained by RAP to both optimal and many random sequences. For the purpose of the comparison we fix the possible operations available for each subassembly, i.e. we always use the NDBG as we describe in Section~\ref{sec-NDBG-query}, which maintains consistency in the returned partitions. We find optimal sequences using an exhaustive search that considers all possible partitions for each subassembly and scores all of them using the simulator (a process which is more than 100 times slower than RAP). As for the random sequences, in which each next step is selected uniformly at random, we find the average score of 50 such sequences. We present the comparison in Figure~\ref{fig:ScoreComparison}.

In Figure~\ref{fig:seqs} we present sequences for two assemblies, comparing two obtained by RAP alongside two with a low robustness score.
%Each sequence is shown from top to bottom, left to right, and scores are only shown for operations that are not robust (i.e., with a score less than 0.95).
For both assemblies, the sequence with the lower score introduces configurations that hinder the successful insertion of parts, with a clear negative impact on robustness. The impact in the \rect example is especially severe, since four more peg-in-hole operations are performed compared to the only (unavoidable) one in the sequence returned by RAP. We observe similar benefits of RAP in the other instances as well.

\begin{comment}
The score in the SIM+NN case is evaluated using the simulator only.
As the score is stochastic, there are small variations between simulator scores in different runs which can result in different sequences. The displayed scores are therefore the average of 50 runs. 
\end{comment}

\section{Conclusions and Future Work} \label{sec-conc}
In this work, we learned a robustness metric by using a physics simulator to provide the ground truth labels.
%The CNN architecture we proposed is a variant of a popular architecture for image-recognition tasks.
A recent study noted the difficulty of such CNNs in capturing geometric properties such as coordinate transforms~\cite{liu2018intriguing}. We conjecture that a special purpose architecture would further improve our results. %Additionally, whether the simulator scores translate to the real world should be validated with experiments.

%In our work we mainly focus on the geometry of subassemblies. However, optimization based on other practical considerations can also be introduced into our framework.

Even after a sequence is fixed, individual motions can be optimized. As we pointed out, within one NDBG cell, one can typically choose from an infinitude of directions. In the current work we choose the middle direction, which is well defined in the one-dimensional motion space. Selecting a good direction in general is in itself an interesting problem, requiring first to define what makes a direction better than others. More generally, one could replace a one-step translation with an arbitrary path, e.g. to rely on compliant motion to further improve robustness.

%Finally, we hope that this work provides a basis for the naturally desirable goal of handling spatial assemblies, where computing robustness efficiently would again be the main challenge (the geometric planner we use already handles this case).

% In this work, we learned the robustness by using a physics simulator to provide the ground truth labels. In principle, unmodeled physical effects such as friction or flexibility of the parts can also be learned, by collecting data from the true system \danny{what is the true system}. To be practical, such an approach will require a much more data-efficient learning paradigm, such as meta-learning~\cite{lake2015human}.

\begin{comment}
many researchers consider only assembly parts in their model and assume them as free-flying objects that can move in the space among surrounding obstacles, thus avoiding the complications of manipulating parts by robots within tight spaces around the assembly.  (from taxonomy paper)
\end{comment}

\bibliographystyle{IEEEtran}
\bibliography{references}

\end{document}

\appendix
\section*{Appendix A. Network Architecture and Training}
For encoders $\text{Enc}_1$ and $\text{Enc}_2$ in simulation we use stride-2 convolutions with a $5\times5$ kernel. We perform 4 convolutions with filter sizes 64, 128, 256, and 512 followed by two fully-connected layers of size 1024. We use LeakyReLU activations with leak 0.2 for all layers. The translation module $T(z_1, z_2)$ consists of one hidden layer of size 1024 with input as the concatenation of $z_1$ and $z_2$ and output of size 1024. For the decoder $\text{Dec}$ in simulation we have a fully connected layer from the input to four fractionally-strided convolutions with filter sizes 256, 128, 64, 3 and stride $\frac{1}{2}$. We have skip connections from every layer in the context encoder $\text{Enc}_2$ to its corresponding layer in the decoder $\text{Dec}$ by concatenation along the filter dimension. 

For real world images, the encoders perform 4 convolutions with filter sizes 32, 16, 16, 8 and strides 1, 2, 1, 2 respectively. All fully connected layers and feature layers are size 100 instead of 1024. The decoder uses fractionally-strided convolutions with filter sizes 16, 16, 32, 3 with strides $\frac{1}{2}$, 1, $\frac{1}{2}$, 1 respectively. For the real world model only, we apply dropout for every fully connected layer with keep probability 0.5, and we tie the weights of $\text{Enc}_1$ and $\text{Enc}_2$. 

We train using the ADAM optimizer with learning rate $10^{-4}$. We train using 3000 videos for reach, 4500 videos for simulated push, 894 videos for sweep, 180 videos for simulated push with real videos, and 135 videos for real push with real videos.

\section*{Appendix B. Ablation Study}
To evaluate that the different loss functions while training our translation model, and the different components for the reward function while performing imitation, we performed ablations by removing these components one by one during model training or policy learning. To understand the importance of the translation cost, we remove cost $\mathcal{L}_{\text{trans}}$, to understand whether features $z_3$ need to be properly aligned we remove model losses $\mathcal{L}_{\text{rec}}$ and $\mathcal{L}_{\text{align}}$. We see that the removal of each of these losses significantly hurts the performance of subsequent imitation. On removing the feature tracking loss $\hat{R}_\text{feat}$ or the image tracking loss $\hat{R}_\text{image}$ we see that overall performance across tasks is worse.

\begin{figure}[h!]
\centering
  \includegraphics[width=0.95\textwidth]{plot/barablation.png} 
 
  \caption{Ablations on model losses and reward functions for the reaching, pushing and pushing with real world demonstrations tasks. Across tasks, all components of the model are necessary for success.}
  \label{fig:allablations}
\end{figure}

\newpage
\section*{Appendix C. Sample Videos}

\subsection*{Reach Simulation}
\begin{figure}[h!]
  \centering
  \includegraphics[width=0.9\textwidth]{diagrams/reach/reach6.jpg}
  \includegraphics[width=0.9\textwidth]{diagrams/reach/reach8.jpg}
  \caption{Example expert training demonstrations from different viewpoints with variations in color, distractor objects, and goal position.}
\end{figure}

\begin{figure}[h!]
  \small{Source Video} \hspace*{0.73cm}
  \vcenteredinclude{
  \includegraphics[width=0.8\textwidth]{diagrams/reach/reachsrc16.jpg}}\\
  \small{Target Context $o_0$} \hspace*{0.45cm}
  \vcenteredinclude{\includegraphics[width=0.145\textwidth]{diagrams/reach/reachctx16.png}}\\
  \small{Translated Video} \hspace*{0.3cm}
  \vcenteredinclude{\includegraphics[width=0.8\textwidth]{diagrams/reach/reachtrans16.jpg}}\\
  
  \small{Source Video} \hspace*{0.73cm}
  \vcenteredinclude{
  \includegraphics[width=0.8\textwidth]{diagrams/reach/reachsrc21.jpg}}\\
  \small{Target Context $o_0$} \hspace*{0.45cm}
  \vcenteredinclude{\includegraphics[width=0.145\textwidth]{diagrams/reach/reachctx21.png}}\\
  \small{Translated Video} \hspace*{0.3cm}
  \vcenteredinclude{\includegraphics[width=0.8\textwidth]{diagrams/reach/reachtrans21.jpg}}\\
  \caption{Example illustrations of demonstrations for a reaching task (top) being performed in a new context (middle), with the translated observation sequences (bottom).}
\end{figure}
\newpage

\subsection*{Push Simulation}
\begin{figure}[h!]
  \centering
  \includegraphics[width=0.9\textwidth]{diagrams/push/push0.jpg}
  \includegraphics[width=0.9\textwidth]{diagrams/push/push6.jpg}
  \caption{Example expert training demonstrations from different viewpoints with variations in distractor objects, start and goal position.}
\end{figure}

\begin{figure}[h!]
  \small{Source Video} \hspace*{0.73cm}
  \vcenteredinclude{
  \includegraphics[width=0.8\textwidth]{diagrams/push/push2src.jpg}}\\
  \small{Target Context $o_0$} \hspace*{0.45cm}
  \vcenteredinclude{\includegraphics[width=0.145\textwidth]{diagrams/push/push2ctx.png}}\\
  \small{Translated Video} \hspace*{0.3cm}
  \vcenteredinclude{\includegraphics[width=0.8\textwidth]{diagrams/push/push2trans.jpg}}\\
  
  \small{Source Video} \hspace*{0.73cm}
  \vcenteredinclude{
  \includegraphics[width=0.8\textwidth]{diagrams/push/push6src.jpg}}\\
  \small{Target Context $o_0$} \hspace*{0.45cm}
  \vcenteredinclude{\includegraphics[width=0.145\textwidth]{diagrams/push/push6ctx.png}}\\
  \small{Translated Video} \hspace*{0.3cm}
  \vcenteredinclude{\includegraphics[width=0.8\textwidth]{diagrams/push/push6trans.jpg}}\\
  \caption{Example illustrations of demonstrations for a pushing task (top) being performed in a new context (middle), with the translated observation sequences (bottom).}
\end{figure}

\newpage
\subsection*{Sweep Simulation}
\begin{figure}[h!]
  \centering
  \includegraphics[width=0.9\textwidth]{diagrams/sweep/sweep5.jpg}
  \includegraphics[width=0.9\textwidth]{diagrams/sweep/sweep7.jpg}
  \caption{Example expert training demonstrations from different viewpoints.}
\end{figure}

\begin{figure}[h]
  \small{Source Video} \hspace*{0.73cm}
  \vcenteredinclude{
  \includegraphics[width=0.8\textwidth]{diagrams/sweep/sweep25src.jpg}}\\
  \small{Target Context $o_0$} \hspace*{0.45cm}
  \vcenteredinclude{\includegraphics[width=0.145\textwidth]{diagrams/sweep/sweep25ctx.png}}\\
  \small{Translated Video} \hspace*{0.3cm}
  \vcenteredinclude{\includegraphics[width=0.8\textwidth]{diagrams/sweep/sweep25trans.jpg}}\\
  
  \small{Source Video} \hspace*{0.73cm}
  \vcenteredinclude{
  \includegraphics[width=0.8\textwidth]{diagrams/sweep/sweep23src.jpg}}\\
  \small{Target Context $o_0$} \hspace*{0.45cm}
  \vcenteredinclude{\includegraphics[width=0.145\textwidth]{diagrams/sweep/sweep23ctx.png}}\\
  \small{Translated Video} \hspace*{0.3cm}
  \vcenteredinclude{\includegraphics[width=0.8\textwidth]{diagrams/sweep/sweep23trans.jpg}}\\
  \caption{Example illustrations of demonstrations for a sweeping task (top) being performed in a new context (middle), with the translated observation sequences (bottom).}
\end{figure}

\newpage
\subsection*{Striking Simulation}
\begin{figure}[h!]
  \centering
  \includegraphics[width=0.8\textwidth]{diagrams/strike/strike1.jpg}
  \includegraphics[width=0.8\textwidth]{diagrams/strike/strike2.jpg}
  \caption{Example expert training demonstrations from different viewpoints.}
\end{figure}

\begin{figure}[h]
  \small{Source Video} \hspace*{0.73cm}
  \vcenteredinclude{
  \includegraphics[width=0.8\textwidth]{diagrams/strike/strikesrc1.jpg}}\\
  \small{Target Context $o_0$} \hspace*{0.45cm}
  \vcenteredinclude{\includegraphics[width=0.145\textwidth]{diagrams/strike/strikectx1.png}}\\
  \small{Translated Video} \hspace*{0.3cm}
  \vcenteredinclude{\includegraphics[width=0.8\textwidth]{diagrams/strike/striketrans1.jpg}}\\
  
  \small{Source Video} \hspace*{0.73cm}
  \vcenteredinclude{
  \includegraphics[width=0.8\textwidth]{diagrams/strike/strikesrc2.jpg}}\\
  \small{Target Context $o_0$} \hspace*{0.45cm}
  \vcenteredinclude{\includegraphics[width=0.145\textwidth]{diagrams/strike/strikectx2.png}}\\
  \small{Translated Video} \hspace*{0.3cm}
  \vcenteredinclude{\includegraphics[width=0.8\textwidth]{diagrams/strike/striketrans2.jpg}}\\
  \caption{Example illustrations of demonstrations for a striking task (top) being performed in a new context (middle), with the translated observation sequences (bottom).}
\end{figure}